\documentclass{llncs}

\usepackage[francais,english]{babel}

\usepackage{listings}
\lstdefinestyle{customc}{
  belowcaptionskip=1\baselineskip,
  breaklines=true,
  frame=L,
  xleftmargin=\parindent,
  language=SQL,
  showstringspaces=false,
  basicstyle=\fontsize{7}{9}\ttfamily,
  keywordstyle=\bfseries\color{black!40!black},
  identifierstyle=\color{black},
  stringstyle=\bfseries\color{Blue!40!Blue},
}

\usepackage{textpos}
\usepackage{blindtext}
\usepackage{placeins}
\usepackage[dvipsnames]{xcolor}
\usepackage{amsthm}
\usepackage{framed}
\usepackage{booktabs}
\usepackage{xspace}

\usepackage{url}
\usepackage[toc,page]{appendix}
\usepackage{multirow,array}

\usepackage{soul}

\usepackage[utf8]{inputenc}

\usepackage{amsfonts,amstext,amsmath,amssymb}
\usepackage{wasysym}

\usepackage{enumerate}

\usepackage{cancel}
\usepackage{caption}
\usepackage{graphicx}
\usepackage{subfig}
\usepackage{float}
\usepackage{graphics}
\usepackage{fancyhdr}
\usepackage{wrapfig}

\graphicspath{{figures/}}

\usepackage{amsmath}
\usepackage[linesnumbered,ruled,vlined]{algorithm2e}
\usepackage[noend]{algcompatible}

\newcommand{\ie}{\textit{i}.\textit{e}., }
\newcommand{\Ie}{\textit{I}.\textit{e}., }
\newcommand{\eg}{\textit{e}.\textit{g}., }

\newcommand{\wrt}{\textit{w}.\textit{r}.\textit{t}.{ }}

\newcommand{\lars}{LARS\xspace}

\usepackage{color}
\usepackage{times}
\usepackage{soul}

\title{BigSR: an empirical study of real-time expressive RDF stream reasoning on modern Big Data platforms}
\author{Xiangnan Ren$^{1,2}$, Olivier Curé$^{2}$, Hubert Naacke$^{3}$, and Guohui Xiao$^{4}$}
\institute{	
 $^{1}$ ATOS, France\\
 \email{xiang-nan.ren@atos.net}\\
 $^{2}$ LIGM (UMR 8049), CNRS, UPEM, France  \\
   \email{olivier.cure@u-pem.fr}\\
$^{3}$ Sorbonne Universit\'es, UPMC, Univ Paris 06, France  \\
   \email{hubert.naacke@lip6.fr}\\
$^{4}$ Free University of Bozen-Bolzano, Italy   \\
   \email{xiao@inf.unibz.it}
}

\begin{document}
\maketitle

\vspace{-5mm}
\begin{abstract}
The trade-off between language expressiveness and system scalability
(E\&S) is a well-known problem in RDF stream reasoning. Higher expressiveness supports more complex reasoning logic, however, it may also hinder system scalability. Current research mainly focuses on logical frameworks suitable for stream reasoning as well as the implementation and the evaluation of prototype systems. These systems are normally developed in a centralized setting which suffer from inherent limited scalability, while an in-depth study of applying distributed solutions to cover E\&S is still missing. In this paper, we aim to explore the feasibility of applying modern distributed computing frameworks to meet E\&S all together. To do so, we first propose BigSR, a technical demonstrator that supports a positive fragment of the LARS framework. For the sake of generality and to cover a wide variety of use cases, BigSR relies on the two main execution models adopted by major distributed execution frameworks: Bulk Synchronous Processing (BSP) and Record-at-A-Time (RAT). Accordingly, we implement BigSR on top of Apache Spark Streaming (BSP model) and Apache Flink (RAT model). In order to conclude on the impacts of BSP and RAT on E\&S, we analyze the ability of the two models to support distributed stream reasoning and identify several types of use cases characterized by their levels of support. This classification allows for quantifying the E\&S trade-off by assessing the scalability of each type of use case \wrt its level of expressiveness. Then, we conduct a series of experiments with 15 queries from 4 different datasets. Our experiments show that BigSR over both BSP and RAT generally scales up to high throughput beyond million-triples per second (with or without recursion), and RAT attains sub-millisecond delay for stateless query operators.

\end{abstract}

%%%%%%%%%%%%%%%%%%%%%%%%%%%%%%%%%%%%%%%%%%%%%%%%%%%%%%%%%%%%%%%%%%%%%%
%%%%%%%%%%%%%%%%%%%%%%%%%%%%%%%%%%%%%%%%%%%%%%%%%%%%%%%%%%%%%%%%%%%%%%
%%%%%%%%%%%%%%%%%%%%%%%    Introduction     %%%%%%%%%%%%%%%%%%%%%%%%%%
%%%%%%%%%%%%%%%%%%%%%%%%%%%%%%%%%%%%%%%%%%%%%%%%%%%%%%%%%%%%%%%%%%%%%%
%%%%%%%%%%%%%%%%%%%%%%%%%%%%%%%%%%%%%%%%%%%%%%%%%%%%%%%%%%%%%%%%%%%%%%

\section{Introduction}
In the era of the ever growing semantic data flood, the challenge of processing declarative queries (or reasoning) over rich and massive RDF data streams remains of major importance. The trade-off between language expressiveness and system scalability (E\&S) has been identified as a main issue in RDF stream reasoning. In the first place, stream processing must be efficient enough to ingest data at the throughput and latency  imposed by the incoming data streams. Besides, the query language has to be expressive enough to support temporal logic and reasoning that may require recursion.  In order to cope with the first aspect, distributed systems supporting fault tolerance, automatic task distribution and recovery are generally involved. Considering the second aspect, Datalog and ASP programs seem to fit efficiently since they represent a good balance between expressive power, safety, performance and usability. Note that considering such an expressiveness permits to address ontology languages such as OWL2RL. This work is not limited to demonstrate that the E\&S trade-off is reachable but we would also like to emphasize that such a solution can be implemented with open-source, state-of-the-art big data technologies, hence being a prototype for production-ready systems.

Thus the BigSR system addresses important problems that are being met frequently in modern applications. For instance projects like Waves\footnote{https://www.waves-rsp.org/} (French FUI), SEAS\footnote{https://www.the-smart-energy.com} (European ITEA2), Optique\footnote{http://optique-project.eu/} (European FP7) and many others require to process data streams with rich semantics in close to real time. At the same time, industrial systems based on Datalog \cite{logiblox}, \cite{yedalog}, datomic\footnote{www.datomic.com}  are emerging. It hence makes sense to mix these two features in a single framework to fulfill an emerging kind of systems.

Available stream reasoning systems \cite{StreamRule,StreamRule_1,ticker} chose a centralized design to benefit from existing ASP solver such as Clingo \cite{Clingo}. The system scalability is apparently limited by single machine/process. Actually, high expressive queries often involve recursion or complex temporal logical operators, which are considered as the main performance bottlenecks for stream reasoning. Additionally, The optimization tailored for static query evaluation \eg data indexing, data preprocessing neither meet the real-time nor the defined temporal logical requirements. And distributed environment often retains shared-nothing context. The memory of different data partitions are isolated, some advanced optimization for Datalog program materialization \cite{FB_Datalog} can be hardly applied.

Although it may have many obstacles to apply distributed computing technique on stream reasoning, relevant in-depth research is still missing. In this paper we first introduce our technical demonstrator, namely BigSR. BigSR possesses two broad classes of streaming models: Bulk Synchronous Processing (BSP) and Record-at-A-Time(RAT) to cover a wide variety of use cases. We implement BigSR on Apache Spark Streaming \cite{Sparkstreaming} and Apache Flink \cite{Flink} to support our evaluations.

Our key contributions are as follows: (1) we build a connection between recent theoretical works on RDF stream reasoning to the state of the art Big Data technologies. We also first try to combine stream reasoning (with complex temporal logics and recursion) with distributed computing. (2) We implement a reusable prototype to support a positive fragment of \lars framework on two distributed systems, namely Apache Spark and Apache Flink. (3) We identify the pros and cons of BSP and RAT on E\&S for different scenarios, respectively. (4) We conduct a series of experiments from variety datasets, and through our experiments we prove that both E\&S can be achieved with distributed solution.

%%%%%%%%%%%%%%%%%%%%%%%%%%%%%%%%%%%%%%%%%%%%%%%%%%%%%%%%%%%%%%%%%%%%%%
%%%%%%%%%%%%%%%%%%%%%%%%%%%%%%%%%%%%%%%%%%%%%%%%%%%%%%%%%%%%%%%%%%%%%%
%%%%%%%%%%%%%%%%%%%%%%     Background Knowledge    %%%%%%%%%%%%%%%%%%%
%%%%%%%%%%%%%%%%%%%%%%%%%%%%%%%%%%%%%%%%%%%%%%%%%%%%%%%%%%%%%%%%%%%%%%
%%%%%%%%%%%%%%%%%%%%%%%%%%%%%%%%%%%%%%%%%%%%%%%%%%%%%%%%%%%%%%%%%%%%%%

\section{Background Knowledge}
We begin by giving the basic notions of \lars. Next, we describe the general methodologies to parallelize the evaluation of  Datalog programs and their relations to \lars. Finally, we introduce two representative distributed streaming systems, \ie Spark Streaming and Flink with their own streaming models BSP and RAT, respectively. We assume the readers are familiar with the basic notions of Datalog, \eg, intensional and extensional databases (resp. IDB and EDB), stratification, recursion \cite{FOD}.

\subsection{LARS Framework}
LARS~\cite{Lars} is a rule-based logical framework defined as an extension of Answer Set Programming (ASP) which we are using as a theoretical foundation. Systems like Ticker\cite{ticker} and Laser\cite{laser} are also based on LARS.
For the sake of briefness, we only list some basic definitions and
the \lars formal syntax that we use in our system. 

Assume an atom set
$\mathcal{A} = \mathcal{A^{\mathcal{I}}} \cup \mathcal{A^{\mathcal{E}}}$, where $\mathcal{A^{\mathcal{I}}}$ is a set of \emph{intensional} atoms and $\mathcal{A^{\mathcal{E}}}$ a set of \emph{extensional} atoms disjoint from $\mathcal{A^{\mathcal{I}}}$. In this paper, a \emph{term} starting with a capital letter refers to a  variable, otherwise it is constant.

\begin{definition}[Stream]
  A stream is a pair $S = (T, v)$, where $T$ is a timeline interval in
  $\mathbb{N}$, and $v: \mathbb{N} \rightarrow 2^{\mathcal{A}}$ is an
  evaluation function such that $v(t)=0$ for
  $t \in \mathbb{N}\setminus T$.

  A stream $S'=(T\,v')$ is a substream of $S = (T, v)$, if
  $T' \subseteq T$, and $v'(t') \subseteq v(t')$ for all $t' \in T'$,.
  \vspace*{-1.5mm}
 \end{definition}

\begin{definition}[Window function]
A window function $w$ takes a stream $S = (T, v)$ as input and returns a substream $S'$, where $S' = (T', v')$.  where $T' \subseteq T$, $\forall t' \in  T'$, $v'(t') \subseteq v(t')$. $S'$ selects the most recent atoms in the $n$ time points. 
\vspace*{-1.5mm}
\end{definition}

\begin{definition}[Time-based Window]
Consider a stream $S = (T, v)$, $T =\mathopen[t_{min},t_{max}\mathclose]$ and a pair $(\ell, d) \in \mathbb{N} \cup \{\infty\}$. A time-based window
$w_\iota(S,t,\ell,u)$ returns the substream $S'$ of $S$ contains all
the elements of the last $\ell$ time units, and $w$ slides with step
size $d$.

\textnormal{\lars distinguishes two types of streams - $S$ and $S^{\star}$. $S$
represents the currently considered window $S$, and $S^{\star}$ is called fixed input stream. To meet the real-time feature, we consider that $S$ as the type of input stream. \Ie we do not assume that the system is capable to load the stream $S = (T, v)$, $T = \mathopen[t_{min}, t_{max}\mathclose]$ from $t_{min}$ to $t_{max}$ directly.}
\vspace*{-1.5mm}
\end{definition} 

\begin{definition}[Window operators]
Let $w$ be a window function. The window operator
$\boxplus^{w}$ signifies that the evaluation should occur on the delivered stream by window function $w$.

\textnormal{We consider the set $\mathcal{A}^+$ of extended atoms by the grammar: $a ~|~ \Diamond \alpha ~|~ \boxplus^w\alpha$ , where $a \in \mathcal{A}$ and $t \in \mathbb{N}$ is a time point. The formula $\Diamond \alpha$ means $\Diamond \alpha$ holds in the current window $S$, if $\alpha$ holds at some time point in $S$. The window operator $\boxplus^{w}$ signifies that the evaluation should occur on the delivered stream by window function $w$.}
\vspace*{-1.5mm}
\end{definition}

\begin{definition}[Rule and program]
An expression of the form $\alpha \leftarrow \beta_1, \ldots, \beta_n$ is called a \lars rule. \noindent{}where $\alpha$ is an atom and $\beta_1 , \ldots \beta_n$
are extended atoms. A (positive plain) \lars program $\mathcal{P}$ is a set of \lars rules.
\vspace*{-1.5mm}
\end{definition} 

The semantics of \lars programs is given by the notion of \emph{answer stream}. For a positive \lars program, its answer stream is unique.

%%%%%%%%%%%%%%%%%%%%%%%%%%%%%%%%%%%%%%%%%%%%%%%%%%%%%%%%%%%%%%%%%%%%%%
%%%%%%%%%%%%%%%%%%%%%%%%%%%%%%%%%%%%%%%%%%%%%%%%%%%%%%%%%%%%%%%%%%%%%%
%%%%%%%%%%%%%%%%%%%%       System Architecture       %%%%%%%%%%%%%%%%%
%%%%%%%%%%%%%%%%%%%%%%%%%%%%%%%%%%%%%%%%%%%%%%%%%%%%%%%%%%%%%%%%%%%%%%
%%%%%%%%%%%%%%%%%%%%%%%%%%%%%%%%%%%%%%%%%%%%%%%%%%%%%%%%%%%%%%%%%%%%%%

\subsection{Parallel Datalog evaluation}
\label{subsec:paraDatalog}
Before illustrating the distributed stream reasoning in BigSR, we summarize the three Parallelism Levels (PL) mentioned in \cite{ParaASP} for parallel instantiation of Datalog programs.

\textbf{(PL1) Components level.} Consider a stratified Datalog
program $\mathcal{P}$ and its dependency graph
$G_{\mathcal{P}} = \langle V,E \rangle$. $\mathcal{P}$ can be split
into $n$ subprograms $\{p_i\}_{i \in 1,...,n}$ where each subprogram $p_i$ is associated to a strongly connected component (SCC) $C_i$ of $G_{\mathcal{P}}$. 
In accordance with the topological order of $C_i$, we can identify the subprograms that can be executed in parallel.

\textbf{(PL2) Rules level.} When recursion occurs in $C_i$, $p_i$ is
evaluated through bottom-up semi-naive algorithm \cite{FOD} be evaluated concurrently.

\textbf{(PL3) Single Rule level.} Consider a program $\mathcal{P} = \mathrm{T}(\mathit{X}) \leftarrow \mathrm{R}(\mathit{Y},\mathit{X})$ where $\mathcal{P}$ contains a limited number of rules. As a result, $\mathcal{P}$ is neither benefiting from PL1 nor PL2. In this situation, the idea is to divide a single rule instantiation into a number of subtasks. All the subtasks are able to run independently.

Computing the answer stream of a positive \lars program can be
regarded as evaluating a Datalog programs at each time point. PL1, PL2 and PL3 are thus enabled to be applied in \lars program.

\subsection{Streaming Models: BSP and RAT}
\textbf{Streaming Models.} In general, two broad classes of execution models exist in distributed stream processing frameworks: BSP\cite{bsp} and RAT. The representative streaming systems with BSP model like Spark Streaming\cite{Spark}, Google Dataflow \cite{DataFlow} buffer and process data by batch. BSP organizes the communication between processes and synchronizes the data processing across records by setting barriers at the end of each batch. On the contrary, RAT systems like Flink \cite{Flink} and Storm \cite{Storm} handle data processing record by record, where operators are regarded as long running tasks, which rely on mutable local states. The computation is done through the data flowing from one operator to another. We choose Spark and Flink as the underlying systems of BigSR, which correspond to BSP and RAT, respectively. Both systems ensure fault tolerance, automatic work distribution and load balancing.

\textbf{Spark} is a MapReduce-like distributed computing framework. The computation is based on Spark's native data structure, namely, Resilient Distributed Dataset (RDD). An RDD is distributed into multiple partitions across different machines and thus enables operations to be performed in parallel. Spark Streaming is one of the main extension of the Spark engine. Spark Streaming processes data streams by cutting input data into batches which are represented as DStream (sequence of RDDs). The online streaming job is treated as a sequence of bounded batch jobs. To simplify, we denote Spark Streaming as Spark in the rest of this paper. 

\textbf{Flink} is another distributed stream processing engine. Different from Spark Streaming, Flink relies on continuous processing model where the data operations are processed over unbounded dataset. Flink continuously processes and produces incoming data stream.

\begin{figure}[h]
\vspace{-3.5mm}
\advance\leftskip-0.25cm
\includegraphics[width=1.05\linewidth]{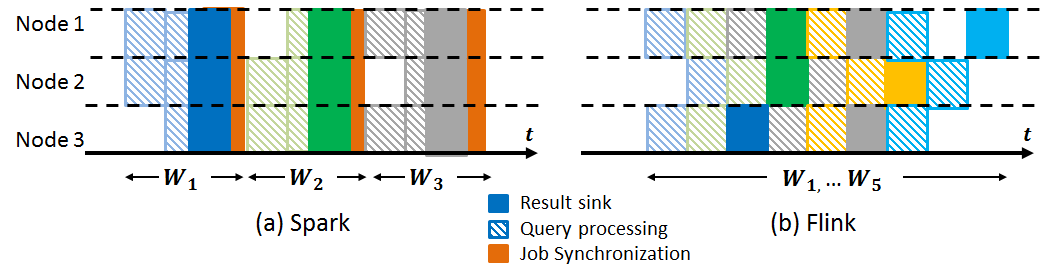}
\captionof{figure}{Blocking and non-blocking query processing.}
\label{fig:asy}
\vspace{-6mm}
\end{figure}

 Before we present the implementation details of BigSR, we use an example to demonstrate a fundamental difference between BSP and RAT. Considering a program 
$\mathcal{P} = \mathrm{T}(\mathit{X}) \leftarrow \boxplus^{w(l,u)}_{\tau}\Diamond(\mathrm{R1}(\mathit{Y}) \wedge \mathrm{R2}(\mathit{Y},\mathit{X}))$,  Figure \ref{fig:asy} roughly gives a runtime example of $\mathcal{P}$ on Spark (Figure \ref{fig:asy}(a)) and Flink (Figure \ref{fig:asy}(b)). Spark processes the data stream synchronously, the next query execution will be launched after the previous one is finished. Conversely, Flink serializes, caches, and pushes forward each record to the next operator eagerly right after the current computation is done. Such behavior minimizes the data processing delay, and operators are able to perform asynchronously. In Figure \ref{fig:asy}, although Spark uses 3 nodes to compute the second part of every query with higher parallelism than Flink, it still needs some  time for synchronization between two jobs. Thus, within 9 time units, Flink is able to finish 5 continuous queries while Spark is only able to process 3 queries during this same period of time. This clearly showcases a better use of computing resources in favor of Flink.

Even \lars dose not include any notion of \emph{state}, to build the connection between \lars and Spark/Flink's execution model, we define stateful and stateless operator as following: A stateless operator over a stream transforms a stream into another stream. In contrast, a statefull operator is a function which takes a pair of a stream and a state, and returns another pair of stream and state. In other words, the data processing in stateless operator only looks up current record. The evaluation of stateful operator requires system to hold an internal state, \eg use local memory or external database for window operator, etc.

\section{Distributed RDF Stream Reasoning}
In this section, we start with a description of the BigSR system architecture. Then, using a query example, we discuss the approaches used to handle distributed RDF stream reasoning based on the BSP and RAT models. Finally, some implementation details on Spark and Flink are provided.

\subsection{Architecture of BigSR}
\begin{figure}[h]
\vspace{-7mm}
\includegraphics[width=1\linewidth]{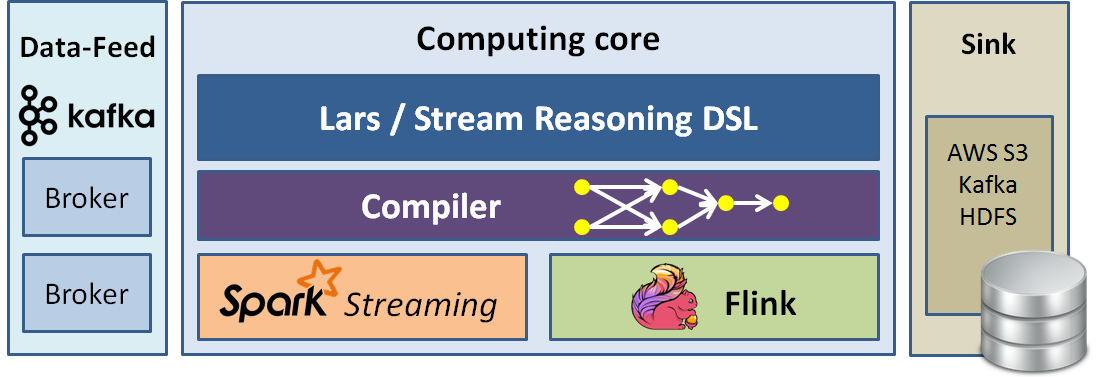}
\captionof{figure}{BigSR system architecture}
\label{fig:bigsr}
\vspace{-5mm}
\end{figure}

Figure \ref{fig:bigsr} gives a high-level view of the BigSR architecture. It consists of three principal modules: (i) \emph{Data-feed} is built on top of Apache Kafka (a distributed message queue) and ensures high throughput and fault-tolerant data stream injection/management; (ii) \emph{Sink} persists query outputs into a storage component such as Amazon S3, HDFS or even Kafka; (iii) \emph{Computing core} first registers and compiles a given \lars program into BigSR's logical execution plan. Then, the system binds the obtained logical plan to the physical operators of Spark (Streaming) or Flink for real-time distributed RDF stream reasoning. 

\begin{center}
\vspace{-3mm}
\begin{lstlisting}[frame=single,caption=BigSR DSL code snippet,label=exampleDSL,captionpos=b]
val atom_tp1 = Atom(procedure, Term("Obs"), Term("Sen"))
val atom_tp2 = Atom(type, Term("Obs"), Term("rainObs"))
val atom_res = Atom(resIRI, Term("Obs"), Term("Sen"))
val r = Rule(atom_res, Set(atom_tp1, atom_tp2), timeWindow(l, u))
val p = Program(Set(r))
\end{lstlisting}
\vspace{-7mm}
\end{center}

BigSR comes with a \lars/stream reasoning DSL language. Listing \ref{exampleDSL}
showcases a query example from the SRBench dataset. The query grammar follows a general datalog program writing style. For instance on line 2 of Listing \ref{exampleDSL}, \textsf{atom\_tp2} denotes an atom of extensional predicate $type$ with variable $Obs$ and constant $rainObs$. Rule \textsf{r} is constructed
with a head atom \textsf{atom\_res}, two body atoms \textsf{atom\_1},
\textsf{atom\_2} and a time-based window
\textsf{timeWindow}. \textsf{timeWindow} accepts two parameters
\textsf{l} and \textsf{u} to define a sliding window over the
conjunction of \textsf{atom\_tp1} and \textsf{atom\_tp2}. Finally, we
construct program \textsf{p}, where \textsf{p} can be expressed by
the following \lars rule:
\begin{center}
  \vspace*{-2mm}
  $\mathrm{resIRI}(\mathit{Obs}, \mathit{Sen}) \leftarrow
  \boxplus^{w(l,u)}_{\tau}\Diamond(\mathrm{procedure}(\mathit{Obs},
  \mathit{Sen}) \wedge \mathrm{type}(\mathit{Obs}, \mathit{rainObs}).$
\end{center}

\subsection{ Data Structure \& Program plans generation}
In order to capture the previously-stated parallelism paradigms of Section \ref{subsec:paraDatalog} with BSP and RAT streaming models, we detail the query evaluation on Spark and Flink. BigSR adopts set semantics to handle all stateful operators, \ie each IDB inferred by stateful datalog formulas will be deduplicated. 

\textbf{Data Structure.} Both Spark and Flink keep their own data structures to support BSP and RAT, respectively: (1) Spark abstracts a sequence of RDD as \emph{DStream}\cite{dstreams} to enable near real-time data processing, the system buffers incoming data streams periodically as a micro-batch RDD. Each RDD encapsulates the data in a certain time interval, \ie corresponding to wall-clock times. 
Intuitively, a micro-batch RDD refers to the minimum allowable data operation granularity. In addition, the timestamp assigned by the system does not bring any impact to query's semantic. (2) Flink takes \emph{DataStream} as a basic data structure. DataStream represents a parallel data flow running on multiple partitions where all data transformations are processed at a record-level. Such fine-grained data transformation makes event-timestamp-based operation feasible. In BigSR, we use the so-called \emph{ingestion time} to handle the time-based window. Practically, each record gets stream source's current time as a timestamp. Moreover, internally, Flink handles ingestion and event time in the same manner. Instead of using event timestamp, we choose ingestion time for data stream processing in Flink for the purpose of simplifying our experiment. 

\begin{figure}[h]
\vspace{-5mm}
\advance\leftskip-0.1cm
\includegraphics[width=1.05\linewidth]{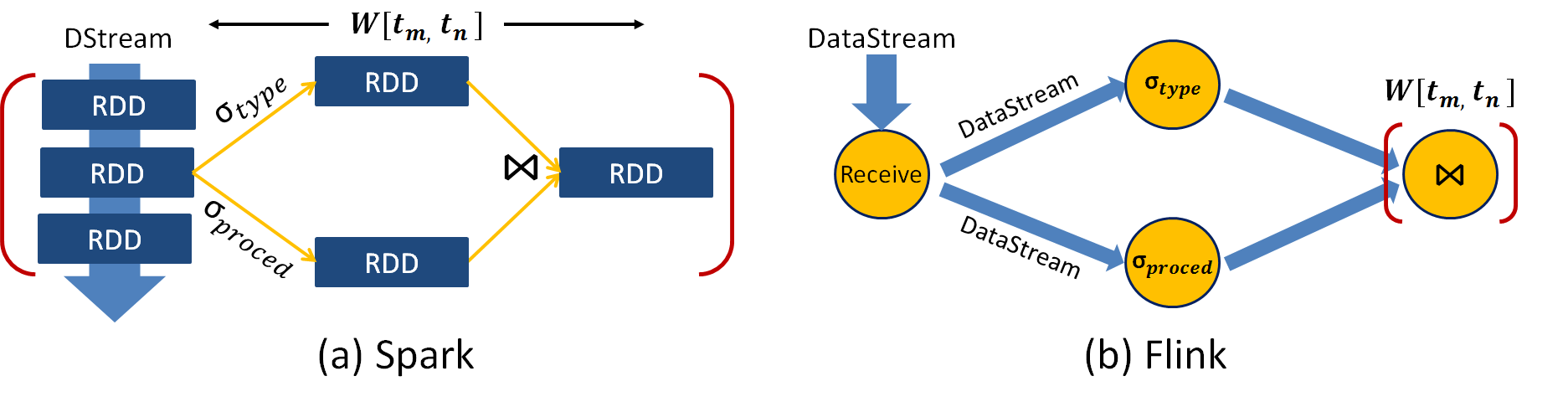}
\captionof{figure}{Logical plan of query \ref{exampleDSL} on Spark and Flink}
\label{fig:dag}
\vspace{-7mm}
\end{figure}

\textbf{Program plans generation.} Figure \ref{fig:dag} compares the logical plan of example query in Listing \ref{exampleDSL}. Both Spark and Flink can naturally embed the three parallelism strategies from Section \ref{subsec:paraDatalog} into their own native physical plan. Due to reliable cluster resource allocation, continuous Spark jobs are launched synchronously, \ie a link connects two consecutive query executions (Figure \ref{fig:dag}, (a)). Input data from stream sources are collected through a window of duration $T$ which consists of $n$ micro-batches of duration $t$. The buffered data are processed by the entire DAG of operators. On the other hand, the execution bound only exists on stateful operators, \eg join and window, in Flink (Figure \ref{fig:dag} (b)). Except for the conjunction of $\sigma_{type}$ and $\sigma_{procedure}$ which run synchronously, data reception and selections $\sigma_{type}$, $\sigma_{procedure}$ run asynchronously.

\subsection{Distributed Stream Reasoning on Spark}
In order to clarify distributed stream reasoning in Spark, we introduce the following program ($\mathcal{P}$) where we assume that $R1$, $R2$, and $R3$ hold the same window operator in their bodies. Predicate $\mathrm{p}_0$ is an EDB predicate while  $\mathrm{p}_1$ and $\mathrm{p}_2$ are two IDB predicates.

% \vspace*{-5mm}
\begin{align*}  
&R1: \hskip 1em &\mathrm{p}_2(\mathit{X}, \mathit{Y}) \leftarrow &\boxplus^{w(l,u)}_{\tau}\Diamond(\mathrm{p}_0(X, Y)) \\
&R2: \hskip 1em &\mathrm{p}_1(\mathit{X}, \mathit{Y}) \leftarrow &\boxplus^{w(l,u)}_{\tau}\Diamond(\mathrm{p}_2(X, Y) \wedge \mathrm{p}_0(\mathit{Y}, \mathit{Z})) \\
&R3: \hskip 1em &\mathrm{p}_2(\mathit{X}, \mathit{Y}) \leftarrow &\boxplus^{w(l,u)}_{\tau}\Diamond(\mathrm{p}_1(Y) \wedge \mathrm{p}_0(\mathit{X}, \mathit{Y}))
\end{align*}
% \vspace*{-5mm}

$\mathcal{P}$ corresponds to a series
of BSP Spark \emph{jobs} which execute in a Spark Streaming
context. One restriction is that $\mathcal{P}$ is only allowed to possess a single global window ($\boxplus^{w(l,u)}_{\tau}$). 

\begin{figure}[h]
\vspace{-3mm}
% \advance\leftskip-0.25cm
\includegraphics[width=1 \linewidth]{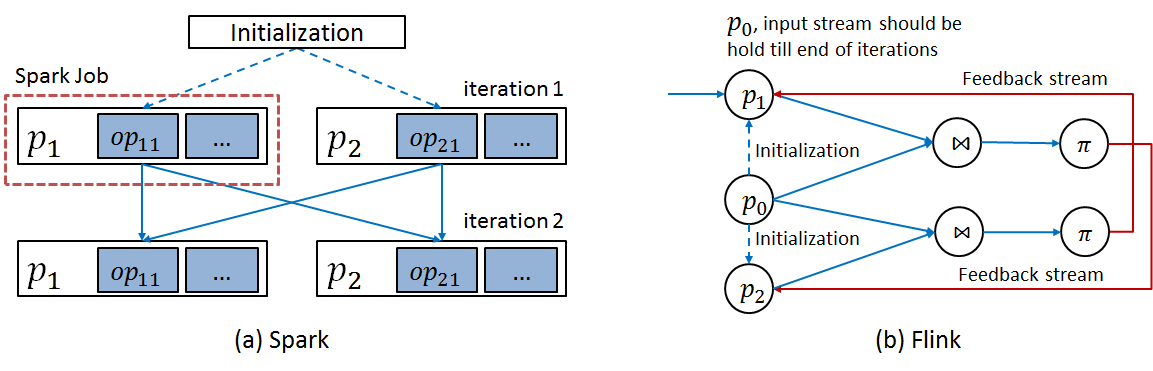}
\captionof{figure}{Recursive query processing on Spark and Flink.}
\label{fig:recursion}
\vspace{-5mm}
\end{figure}

For a non-recursive program, the logical plan is first mapped into a single Spark job's DAG logical plan. Next, Spark compiles the logical plan to its physical plan and is then  evaluated. We mainly discuss the recursive program evaluation on Spark here. 
Considering the previously defined program $\mathcal{P}$, Figure \ref{fig:recursion} (a) gives a running example by using the semi-naive evaluation of \cite{FOD}. We now present how this evaluation is parallelized in Spark:

\vspace*{-3mm}
\begin{itemize}
\item \textbf{PL1.}: The system starts the evaluation by initializing $\mathrm{p}_1$ and $\mathrm{p}_2$ using input EDB $\mathrm{p}_0$. $\mathrm{p}_1$ and $\mathrm{p}_2$ forms an SCC $C_i$, for any other SCC $C_j$, where $C_j$ does not depend on $C_i$, the evaluations of $C_i$ and $C_j$ can be computed in parallel. 

\item \textbf{PL2.} In each iteration step of the semi-naive evaluation, the set of operations (\eg selection, join, union) which compute each IDB predicate are chained together as a Spark job, \ie, two Spark jobs for the evaluation of $\mathrm{p}_1$ and $\mathrm{p}_2$ are involved. The iterations are completed until a fix-point is reached. The system outputs $\mathrm{p}_1$ and $\mathrm{p}_2$ for further calculation.

\item \textbf{PL3.} Inside a single Spark Job, each operator performs a transformation of RDD. As an RDD is a distributed data collection, multiple tasks may execute concurrently across different data partitions.
\vspace{-3mm}
\end{itemize}

% \begin{figure}[h]
% % \vspace{-5mm}
% % \advance\leftskip-0.25cm
% \includegraphics[width=1 \linewidth]{recursion.png}
% \captionof{figure}{Recursive query processing on Spark and Flink.}
% \label{fig:recursion}
% % \vspace{-6mm}
% \end{figure}

\textbf{Limitation \& Envisioned Optimization.} The implementation of recursion support in BigSR is straightforward. Some discussions about \cite{BigDatalog} and our envisioned optimization for recursive query handling are worth mentioning. Of the four proposed solutions in \cite{BigDatalog}, two of them play key roles: (i) extends immutable RDD to mutable \emph{SetRDD}; (ii) add \emph{recursive stage} support in Spark job scheduler. Both (i) and (ii) sacrifice fault-tolerance to gain system performance. For a streaming service runs in $24 \times 7$, such fault-tolerance trade-off has high potential impact on system's robustness and reliability. 

In addition to the optimizations of \cite{BigDatalog}, we propose two other envisioned solutions: (a) Integration with distributed index. For stratified \lars program, iterations and join operations become the performance bottleneck, Since iteration potentially involves intensive job scheduling and network shuffling in a distributed environment. Instead of scanning each partition entirely in RDD, the work\footnote{\url{https://github.com/amplab/spark-indexedrdd}} integrates adaptive radix tree \cite{PART} to have efficient data access and update. Our preliminary experiments shows that indexed RDD gains 2.5 - 3 times performance for each iteration. However, the index needs to be reconstructed after each iteration, which comes with a non-negligible overhead. This approach thus is in the experimental and envisaged stage. (b) Parallel job scheduling. Spark schedules submitted jobs in a centralized way by master node. Frequent job submissions in iteration causes unavoidable latency. Rather than merging multiple jobs together into a single job \cite{BigDatalog}, distributed job scheduler \cite{Sparrow} provides millisecond latency for job scheduling without sacrificing any fault-tolerance.

\subsection{Distributed Stream Reasoning on Flink}

For a non-recursive program, BigSR compiles a \lars program
$\mathcal{P}$ into Flink's streaming topology (\ie DAG). Then,
$\mathcal{P}$ is evaluated through data flowing between operators in
the DAG. In the case of recursive program ($\mathcal{P}$), Figure \ref{fig:recursion}(b) gives a high-level vision of Flink's workflow:

\begin{itemize}
\item \textbf{PL1.} Similar to Spark, any other SCC which is independent from $C_i$ can be computed in parallel.

\item \textbf{PL2.} Different from Spark, which splits the recursion into a series of independent jobs. Flink achieves recursion with streaming feedback. In iteration $i$, the system needs to feed back $S^i_f(\mathrm{p}_1)$ and $S^i_f(\mathrm{p}_2)$ (downstream for computing $\mathrm{p}_1$ and $\mathrm{p}_2$, respectively) as the input for iteration $i+1$. And the processes on $S^i_f(\mathrm{p}_1)$ and $S^i_f(\mathrm{p}_2)$ occur in parallel in a single iteration step.

\item \textbf{PL3.} Similar to Spark, the DataStream flows through each operator across  in parallel, each operator consists of
multiple tasks (over some partitions) which can be performed
concurrently (\wrt PL3).
\end{itemize}

\textbf{Limitation \& Envisioned Optimization.} There are two main limitations we found by using the RAT model:  

(1) We require that the conjunction of two atoms $a_1$, $a_2$ should share the same window operator, \eg, a formula
$\alpha = \boxplus^{w_1} \Diamond a_1 \wedge \boxplus^{w_2} \Diamond a_2 $ currently imposes that $w = w_1 = w_2$ ($\ie \alpha =  \boxplus^{w} \Diamond (a_1 \wedge a_2)$). To our experience, this limitation shows up when input stream is type $S$ and the underlying process relies multi-cores/distributed environment. The main difficulties come from synchronization of clocks, task progress, and window trigger mechanism. 
However, a single-core/centralized system with input stream $S^{\star}$ does not suffer from such a synchronization problem. Since all the computations are done sequentially, the program evaluation performs in a quasi-static way without considering the fast updating of $S^{\star}$. 
To the best of our knowledge, CQELSCloud, which is the only implementation with distributed setting, has a similar semantic. Nevertheless, to avoid above mentioned problems, CQELSCloud forbids event timestamp and ``sliding" mechanisms in their window operators. 
The record is emitted eagerly right when the computation is done. 

(2) We do not support recursive queries in Flink yet. Within the \lars framework, the implementation of recursion with RAT could be quite challenging. The example given earlier skips window operator. Once the body atoms of $\mathrm{p}_1$ and $\mathrm{p}_2$ are restricted by temporal logical operator (\eg window operator), the synchronization problem mentioned in (1) will reappear. A naive solution is that the system merges and pushes the input stream by batch into a single operator for the evaluation of recursion, which behaves like a BSP model. However, this approach is against the original intention of using RAT model. The system becomes difficult to scale up since it does not provide any benefice from the asynchronous executions of the operators.  

\subsection{Discussions}

\vspace*{-5mm}
\begin{table}[h!]
\centering
\begin{tabular}{ccc}
\toprule
\textbf{Model} & \textbf{Expressiveness} & \textbf{Recursion} \\
\midrule
\textbf{BSP (Spark)} { }{ }{ }{ }{ }  & { }{ }{ }{ }{ } Low, coarse-grained  { }{ }{ }{ }{ } 
             &{ }{ }{ }{ }{ } Implementation easy { }{ }{ }{ }{ } \\
\midrule
\textbf{RAT (Flink)} { }{ }{ }{ }{ }  & { }{ }{ }{ }{ } High, fine-grained { }{ }{ }{ }{ }
           & { }{ }{ }{ }{ } Implementation difficult { }{ }{ }{ }{ } \\
\bottomrule
\end{tabular}
      \caption{Intuitive comparison between BSP and RAT}
        \label{tab:bsp_rat}
\end{table}
\vspace*{-7mm}

Table \ref{tab:bsp_rat} briefly compares BSP and RAT for implementing a stream reasoning framework like \lars. With BSP on Spark, the manipulation of temporal logical operators is rather coarse-grained with low expressiveness. The query should be evaluated in batches by a global window. Furthermore, the semantic of data timestamps does not influence the semantic of the query. The combination of window operators is not flexible. However, the query evaluation of the BSP model in a window is practically similar to a static data processing. Therefore, BSP greatly simplifies the implementation of recursion. On the other side, RAT model handles data processing record by record. The evaluation of operators can be performed asynchronously. RAT enables to manipulate the timestamp and window operators in a fine-grained fashion. Different types of time (\eg event time, ingestion time, system processing time) can be integrated on RAT. The combination of multiple window operators can be chained together and ran independently. For recursive query evaluation, RAT suffers from synchronization of processes over multiple streams and multiple windows. This makes the implementation of recursion within the \lars framework a challenging problem.

%%%%%%%%%%%%%%%%%%%%%%%%%%%%%%%%%%%%%%%%%%%%%%%%%%%%%%%%%%%%%%%%%%%%%%
%%%%%%%%%%%%%%%%%%%%%%%%%%%%%%%%%%%%%%%%%%%%%%%%%%%%%%%%%%%%%%%%%%%%%%
%%%%%%%%%%%%%%%%%%%%%%%       Evaluation       %%%%%%%%%%%%%%%%%%%%%%%
%%%%%%%%%%%%%%%%%%%%%%%%%%%%%%%%%%%%%%%%%%%%%%%%%%%%%%%%%%%%%%%%%%%%%%
%%%%%%%%%%%%%%%%%%%%%%%%%%%%%%%%%%%%%%%%%%%%%%%%%%%%%%%%%%%%%%%%%%%%%%

\section{Experiments}
The code base (written using the Scala programming language), data sources and test queries are available on GitHub\footnote{\url{https://github.com/renxiangnan/bigsr}}. We conduct our experiments on a Amazon EMR cluster with a Yarn resource manager. The EMR cluster consists of a total of 9 nodes of type m4.xlarge. One node is setup for the Kafka broker and message producer, one node for Apache Zookeeper, seven nodes for Spark/Flink application (one master node and six worker nodes). Each node has 4 CPU virtual cores of 2.4 GHz Intel Xeon E5-2676 v3 processors, 16 GB RAM and 750 MB/s bandwidth. We use Spark 2.2.1, Flink 1.4.0 (broadcast join is disabled), Scala 2.11.7 and Java 8 as evaluation baselines. 
%Hub: I replaced "broadcast hash join" by "broadcast join"

\subsection{Benchmark Design}
\label{subsec:Benchmark}

\textbf{Dataset \& Queries.} Our evaluation is based on synthetic and real-world datasets which involve 4 different datasets and 15 queries (Table \ref{tab:queries}). The 4 datasets correspond to Waves, SRBench \cite{SRBench}, CityBench \cite{City} and LUBM \cite{Lubm}. All the data captured by the Waves, SRBench, and CityBench datasets come from real-world IoT sensors. The Waves dataset describes measures of a potable water network, \eg values of flow, pressure and chlorine levels, etc. SRBench, one of the first RSP benchmark, contains USA weather observations ranging from 2001 to 2009. CityBench simulates a smart city context for RSP applications and concerns sensor measures on  vehicle traffic, parking lot utilization and user location use cases. All the aforementioned datasets come from RSP contexts. It is hard to design a recursive query, because the generated RDF data streams are usually directly converted from flat data (CSV) with few references between entities. Therefore, we use LUBM for recursive query evaluations. 
Query $Q_1$ to $Q_{11}$ include stateful operators for windowing and recursion, where $Q_{12}$ to $Q_{15}$ only contain stateless operators (\eg selection, filter, projection). We evaluate $Q_{12}$ to $Q_{15}$ to highlight the engine performance with BSP and RAT model for low-latency use cases, such as reactive applications.

\vspace*{-3mm}
\begin{table}[h!]
\centering
\begin{tabular}{cccccc}
\toprule
\textbf{} & \textbf{$Q_1$ - $Q_3$} & \textbf{$Q_4$ - $Q_6$} & \textbf{$Q_7$ - $Q_8$} & \textbf{$Q_9$ - $Q_{10}$}& \textbf{$Q_{11}$ - $Q_{15}$} \\
\midrule
\textbf{Dataset} & { }{ } Waves  { }{ } 
             &{ }{ }{ }{ } SRBench { }{ }{ }{ }
             &{ }{ }{ }{ } CityBench { }{ }{ }{ }
             &{ }{ }{ }{ } LUBM { }{ }{ }{ }
             &{ }{ }{ }{ } Synthetic{ }{ }{ }{ } \\
\midrule
\textbf{Recursive} & { }{ }{ }{ } Non { }{ }{ }{ }
             &{ }{ }{ }{ } No { }{ }{ }{ } 
             &{ }{ }{ }{ } No { }{ }{ }{ }
             &{ }{ }{ }{ } Yes { }{ }{ }{ } 
             &{ }{ }{ }{ } No { }{ }{ }{ }\\
\bottomrule
\end{tabular}
      \caption{Test queries and datasets.}
        \label{tab:queries}
\end{table}
\vspace*{-7mm}

To guarantee query semantics consistency between BSP and RAT, the result sets of a given query should be the same on the two distributed models. Input data streams are generated by Kafka message producer, and injected into BigSR in parallel. The average stream rate is around 250,000 to 300,000 triples/second.

\textbf{Performance metrics.} Considering \emph{Benchmarking Streaming Computation Engines at Yahoo!}, the well-know benchmark for distributed streaming systems\footnote{https://yahooeng.tumblr.com/post/135321837876/benchmarking-streaming-computation-engines-at}, we take system throughput and query latency as the principal performance criteria. In particular, we categorize the evaluations into two groups: 

\vspace*{-1mm}
\begin{itemize}
\item Group 1: $Q_1$ to $Q_{11}$ (queries with stateful operators). We denote throughput as the number of triples processed per second by the engine (\ie triples/second). Latency corresponds to the duration taken by BigSR between the arrival of an input and the generation of its output.

\item Group 2: $Q_{12}$ to $Q_{15}$ (queries with only stateless operators). We focus on the minimum latency that the engine is able to attain. On Spark, we first reduce the micro-batch size as much as possible, then we record the query latency for completing the process of current micro-batch. On Flink, the latency of a record  $r$ indicates the time difference between the moment $r$ enters the system and the moment $r'$ outputs from the system.
\end{itemize}
\vspace*{-1mm}

\textbf{Performance tuning} is one of the most important step for the deployment of Spark and Flink applications. Based on our previous experience in \cite{Strider}, we list three important factors which bring significant impact on engine performance, \ie parallelism level, memory management, and data serialization. Unfortunately, there is no fixed rule to configure these parameters in an optimal way. The tuning has to be done empirically. Besides, recursion on Spark may generate long RDD lineage in the driver memory which can lead to stack overflow. We thus periodically trigger the local checkpoint of RDD to truncate the RDD lineage.

\subsection{Evaluation Results \& Discussion}

In this section, we present and discuss the evaluation result over the queries presented in Section \ref{subsec:Benchmark}. 
We do not compare BigSR to the state of the art RSP/(Streaming) ASP systems. This is due to our previous work \cite{Strider} which has already shown a much higher performance gain (1 to 2 orders of magnitude) than available RSPs (C-SPARQL and CQELS). Compared to \cite{Strider}, the Spark implementation in BigSR is approximately 30\% less performant. We partially attribute this to the \emph{distinct} operation which satisfies the set semantic. Thus we still consider that the distributed design of BigSR takes a substantial performance advantage on existing centralized RSP engines.

\begin{figure}[h]
\vspace*{-4mm}
\advance\leftskip-0.25cm
\includegraphics[width=1.05\linewidth]{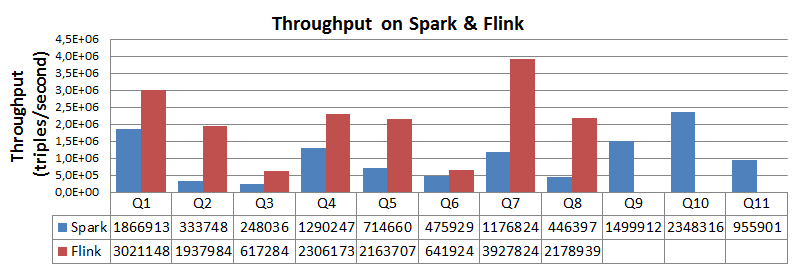}
\captionof{figure}{System throughput (triples/second) on Spark and Flink for $Q_1$ to $Q11$.}
\label{fig:thp_q1_q11}
\vspace*{-2mm}
\end{figure}

\textbf{Throughput.} Figure \ref{fig:thp_q1_q11} reports the engine throughput for $Q_1$ to $Q_{11}$. Both implementation with BSP (Spark) and RAT (Flink) achieves million-level throughput. We observe that the throughput of Flink is 1.x - 3.x times superior to Spark. This difference is more substantial when the query has more intensive joins/conjunctions. 
This can be explained by the job scheduling on Spark which leads join operations to be performed on different compute nodes, thus causing network shuffles. Moreover Spark's job scheduling is difficult to control in a fine-grained manner. On the contrary, Flink is able to avoid shuffles with an appropriate system configuration, \ie the join operation can be managed by a \emph{task manager} and performed locally on a single compute node.
% \hub{Note that a join a operation still benefits from parallel processing (for a multi-cores machine) and that all the compute resources could serve the query under the assumption that the number of operations in the query is greater than the number of compute nodes}

For recursive queries $Q_9$ to $Q_{11}$, Spark achieves the throughput up to 2.3M triples/second. Although we design three recursive queries for Lubm, the length of transitivity in Lubm is small which limits the number of iterations in semi-naive evaluation. Additionally, the intermediate results generated in $Q_9$ to $Q_{11}$ are low-sized, which reduces the performance penalty implied by shuffle operators.

\begin{figure}[h]
\vspace*{-3mm}
\advance\leftskip-0.25cm
\includegraphics[width=1.05\linewidth]{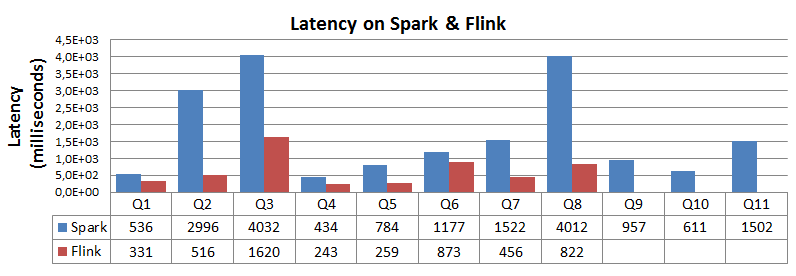}
\captionof{figure}{Query latency (milliseconds) on Spark and Flink for $Q_1$ to $Q11$.}
\label{fig:lat_q1_q11}
\vspace*{-3mm}
\end{figure}

\textbf{Latency.} We summarize the query latency of Group 1 ($Q_1$ to $Q_{11}$) in Figure \ref{fig:lat_q1_q11}. Spark and Flink hold second/sub-second delay in general. Flink has a lower latency than Spark, only $Q_3$ exceeds one second on Flink. The obtained latency on Spark and Flink are already acceptable for most streaming applications. 

Here, we highlight the experiment over queries in group 2 (Table \ref{tab:lat_stateless_query}). 
Intuitively, $Q_{12}$ to $Q_{15}$ have been designed to stress BSP and RAT on latency.
In fact, the micro-batch interval size of Spark is set to 500 ms. Even though the average latency on Spark is around 100 ms, but 500 ms is approximately the minimum ``safe" batch size we can configure on Spark. The reason is that the garbage collection (GC) triggers periodically in a long running Spark Streaming application (on driver and workers), GC pause occurs from time to time. The query latency thus can grow up to 400 ms. We conclude that Spark satisfies the near-real time use case with sub-second delay requirement.

\vspace*{-3mm}
\begin{table}[h!]
\centering
\begin{tabular}{ccccc}
\toprule
\textbf{} & \textbf{$Q_{12}$} & \textbf{$Q_{13}$} & \textbf{$Q_{14}$} & \textbf{$Q_{15}$} \\
\midrule
\textbf{Spark} & { }{ } 110  { }{ } 
             &{ }{ }{ }{ } 96 { }{ }{ }{ }
             &{ }{ }{ }{ } 115 { }{ }{ }{ }
             &{ }{ }{ }{ } 99 { }{ }{ }{ } \\
\midrule
\textbf{Flink} & { }{ }{ }{ } \textless 1 { }{ }{ }{ }
             &{ }{ }{ }{ } \textless 1 { }{ }{ }{ } 
             &{ }{ }{ }{ } \textless 1 { }{ }{ }{ }
             &{ }{ }{ }{ } \textless 1 { }{ }{ }{ } \\
\bottomrule
\end{tabular}
      \caption{Stateless query latency (millisecond); Spark micro-batch size = 500 ms.}
        \label{tab:lat_stateless_query}
\end{table}
\vspace*{-5mm}

On Flink, we calculate the record latency by subtracting the output timestamp from input system assigned timestamp. The minimum observable time unit is millisecond (limited by Flink), and the vast majority obtained latency is 0 ms. Apparently, sub-millisecond delay meets most real-time, latency-sensitive use cases. 

% \vspace*{-3mm}
% \begin{table}[h!]
% \centering
% \begin{tabular}{ccccc}
% \toprule
% \textbf{} & \textbf{$Q_{12}$} & \textbf{$Q_{13}$} & \textbf{$Q_{14}$} & \textbf{$Q_{15}$} \\
% \midrule
% \textbf{Spark} & { }{ } 110  { }{ } 
%              &{ }{ }{ }{ } 96 { }{ }{ }{ }
%              &{ }{ }{ }{ } 115 { }{ }{ }{ }
%              &{ }{ }{ }{ } 99 { }{ }{ }{ } \\
% \midrule
% \textbf{Flink} & { }{ }{ }{ } \textless 1 { }{ }{ }{ }
%              &{ }{ }{ }{ } \textless 1 { }{ }{ }{ } 
%              &{ }{ }{ }{ } \textless 1 { }{ }{ }{ }
%              &{ }{ }{ }{ } \textless 1 { }{ }{ }{ } \\
% \bottomrule
% \end{tabular}
%       \caption{Stateless query latency (millisecond); Spark micro-batch size = 500 ms.}
%         \label{tab:lat_stateless_query}
% \end{table}
% \vspace*{-3mm}

\section{Related Work}
In this section, we consider related work in the context of RSP, stream reasoning and Datalog engines. We emphasize that none of the existing systems tackle both scalability and ASP/Datalog reasoning.

\textbf{RSP.} Several RSP systems have been implemented over the last few years. The
most popular ones correspond to centralized engines, \eg C-SPARQL \cite{CSPARQL}, CQELS \cite{CQELS}, ETALIS \cite{ETALIS} and SPARQL$_{Stream}$ \cite{sparqlStream}. Systems like CQELS-cloud\cite{CQELSCloud} and Strider \cite{Strider} tackle the
scalability issue but distributing stream processing. 
Available RSP systems are equipped with their own syntax, which generally take the form of continuous versions of the standard SPARQL grammar. This limits the expressiveness on temporal logical operators, \eg the combination or even nesting of window operators. Moreover, the support of recursion is also missing.

\textbf{RDF Stream Reasoner.} Both StreamRule \cite{StreamRule} and its recent parallelized version StreamRule${}^P$ \cite{StreamRule_1} use a RSP engine for data stream pre-filtering and Clingo as the ASP solver. The expressiveness of BSP implementation in BigSR can fully cover StreamRule and StreamRule${}^P$, since the implementation in these two reasoners stay on positive stratified Datalog program. Evaluation of StreamRule/StreamRule${}^P$ showcases that the  average throughput is around thousand-triples/second with second-level delay. Comparatively, our BSP implementation has almost three orders of magnitude gains. Laser \cite{laser} and Ticker \cite{ticker} are both stream processing systems based on the LARS framework but do not concentrate on scalability. Ticker concentrates incremental model maintenance and sacrifices performance by relying on an external ASP engine (Clingo). Laser also proposes an incremental model based on time interval annotations which can prevent unnecessary re-computations. Although Laser claims to represent a trade-off between expressiveness and data throughput, it cannot scale the way BigSR enables to. This is mainly due to Laser's inability to distribute stream processing.

\textbf{Other Datalog Solvers.}
Logiblox \cite{logiblox} is a single-machine commercial transactional and analytical system. Its query language, namely LogiQL, is a unified and declarative query language based on Datalog equiped with incremental maintenance.   
  RDFox \cite{RDFox} is a centralized, main-memory RDF store with support for parallel Datalog reasoning and incremental materialization maintenance. None of these systems consider stream processing.
 
Myria \cite{myria} and BigDatalog \cite{BigDatalog} are both distributed datalog engines that perform on shared-nothing architectures. The former is implemented on its parallel processing framework and interacts with PostgreSQL databases for write and read operations. Much of the effort in the datalog engine of Myria has been concentrated on distributing rule processing in a fault-tolerant manner. BigDatalog implements a parallel semi-Naïve datalog evaluation on top of Spark. Neither Myria nor BigDatalog support stream processing.
 
\section{Conclusion}
This paper bridges the gap between theoretical work in progress on RDF stream reasoning and modern cutting-edge Big Data technologies. We emphasize that a trade-off between expressiveness of reasoning and scalability is possible in RDF stream reasoning. In fact our BigSR system is able to reach the millions triples per second processing mark on complex queries and second and subsecond latency in general. In order to tackle scalability, BigSR considers the standard BSP and RAT approaches through implementations with state of the art open source frameworks, respectively Apache Spark and Apache Flink. Both these systems offer rich APIs (\eg obviously for stream processing but also for machine learning, graph analytics), fault-tolerance, load balancing and automatic work distribution. In terms of reasoning, we address logic programming through the ASP-based LARS framework.

Our experimentation presents some interesting results on the current  state of these systems. With its large programmer community, Spark is easier than Flink to get into and implement applications. Nevertheless, it may be difficult to configure, tune this parallel computing framework. The support for recursive rules was not a difficult problem. The overall performance of Flink on both data throughput and latency is superior to Spark and is quite impressive without requiring a lot of tuning. Nonetheless, the design and implementation of an evaluation approach for recursive programs is not straightforward. This is in fact in our future work list together with a more efficient incremental model maintenance. 

%\tiny
\bibliographystyle{plain}
\bibliography{main}

\begin{thebibliography}{10}

\bibitem{FOD}
Serge Abiteboul, Richard Hull, and Victor Vianu.
\newblock {\em Foundations of Databases}.
\newblock Addison-Wesley, 1995.

\bibitem{DataFlow}
Tyler Akidau, Robert Bradshaw, Craig Chambers, Slava Chernyak, Rafael~J.
  Fern\'{a}ndez-Moctezuma, Reuven Lax, Sam McVeety, Daniel Mills, Frances
  Perry, Eric Schmidt, and Sam Whittle.
\newblock The dataflow model: A practical approach to balancing correctness,
  latency, and cost in massive-scale, unbounded, out-of-order data processing.
\newblock {\em PVLDB}, 2015.

\bibitem{City}
Muhammad~Intizar Ali, Feng Gao, and Alessandra Mileo.
\newblock Citybench: A configurable benchmark to evaluate rsp engines using
  smart city datasets.
\newblock In {\em ISWC}, 2015.

\bibitem{ETALIS}
Darko Anicic, Sebastian Rudolph, Paul Fodor, and Nenad Stojanovic.
\newblock Stream reasoning and complex event processing in etalis.
\newblock {\em Semant. web}, 2012.

\bibitem{logiblox}
Molham Aref, Balder ten Cate, Todd~J. Green, Benny Kimelfeld, Dan Olteanu, Emir
  Pasalic, Todd~L. Veldhuizen, and Geoffrey Washburn.
\newblock Design and implementation of the logicblox system.
\newblock In {\em SIGMOD}, 2015.

\bibitem{CSPARQL}
Davide~Francesco Barbieri and al.
\newblock {C-SPARQL:} {SPARQL} for continuous querying.
\newblock In {\em WWW}, 2009.

\bibitem{laser}
Hamid~R. Bazoobandi, Harald Beck, and Jacopo Urbani.
\newblock Expressive stream reasoning with laser.
\newblock In {\em ISWC}, 2017.

\bibitem{Lars}
Harald Beck, Minh Dao{-}Tran, Thomas Eiter, and Michael Fink.
\newblock {LARS:} {A} logic-based framework for analyzing reasoning over
  streams.
\newblock In {\em AAAI}, 2015.

\bibitem{ticker}
Harald Beck, Thomas Eiter, and Christian Folie.
\newblock Ticker: {A} system for incremental asp-based stream reasoning.
\newblock {\em {TPLP}}, 2017.

\bibitem{sparqlStream}
Jean{-}Paul Calbimonte, {\'{O}}scar Corcho, and Alasdair J.~G. Gray.
\newblock Enabling ontology-based access to streaming data sources.
\newblock In {\em ISWC}, 2010.

\bibitem{Flink}
Paris Carbone, Asterios Katsifodimos, Stephan Ewen, Volker Markl, Seif Haridi,
  and Kostas Tzoumas.
\newblock Apache flink{\texttrademark}: Stream and batch processing in a single
  engine.
\newblock {\em {IEEE} Data Eng. Bull.}, 2015.

\bibitem{yedalog}
Brian Chin, Daniel von Dincklage, Vuk Ercegovac, Peter Hawkins, Mark~S. Miller,
  Franz Och, Chris Olston, and Fernando Pereira.
\newblock Yedalog: Exploring knowledge at scale.
\newblock In {\em SNAPL}, 2015.

\bibitem{Clingo}
Martin Gebser, Roland Kaminski, Benjamin Kaufmann, and Torsten Schaub.
\newblock Clingo = {ASP} + control: Preliminary report.
\newblock {\em CoRR}, 2014.

\bibitem{Lubm}
Yuanbo Guo, Zhengxiang Pan, and Jeff Heflin.
\newblock {LUBM:} {A} benchmark for {OWL} knowledge base systems.
\newblock {\em J. Web Sem.}, 2005.

\bibitem{CQELS}
Danh Le-Phuoc, Minh Dao-Tran, Josiane~Xavier Parreira, and Manfred Hauswirth.
\newblock A native and adaptive approach for unified processing of linked
  streams and linked data.
\newblock In {\em ISWC}, 2011.

\bibitem{PART}
Viktor Leis, Alfons Kemper, and Thomas Neumann.
\newblock The adaptive radix tree: Artful indexing for main-memory databases.
\newblock ICDE '13, 2013.

\bibitem{StreamRule}
Alessandra Mileo, Ahmed Abdelrahman, Sean Policarpio, and Manfred Hauswirth.
\newblock Streamrule: {A} nonmonotonic stream reasoning system for the semantic
  web.
\newblock In {\em {RR}}, 2013.

\bibitem{FB_Datalog}
Boris Motik, Yavor Nenov, Robert Piro, Ian Horrocks, and Dan Olteanu.
\newblock Parallel materialisation of datalog programs in centralised,
  main-memory {RDF} systems.
\newblock In {\em AAAI}, 2014.

\bibitem{RDFox}
Boris Motik, Yavor Nenov, Robert Piro, Ian Horrocks, and Dan Olteanu.
\newblock Parallel materialisation of datalog programs in centralised,
  main-memory {RDF} systems.
\newblock In {\em AAAI}, 2014.

\bibitem{Sparrow}
Kay Ousterhout, Patrick Wendell, Matei Zaharia, and Ion Stoica.
\newblock Sparrow: Distributed, low latency scheduling.
\newblock SOSP, 2013.

\bibitem{ParaASP}
Simona Perri, Francesco Ricca, and Marco Sirianni.
\newblock Parallel instantiation of {ASP} programs: techniques and experiments.
\newblock {\em {TPLP}}, 2013.

\bibitem{StreamRule_1}
Thu{-}Le Pham, Alessandra Mileo, and Muhammad~Intizar Ali.
\newblock Towards scalable non-monotonic stream reasoning via input dependency
  analysis.
\newblock In {\em 33rd {ICDE} 2017}, 2017.

\bibitem{CQELSCloud}
Danh~Le Phuoc and al.
\newblock Elastic and scalable processing of linked stream data in the cloud.
\newblock In {\em ISWC}, 2013.

\bibitem{Strider}
Xiangnan Ren and Olivier Cur{\'{e}}.
\newblock Strider: {A} hybrid adaptive distributed {RDF} stream processing
  engine.
\newblock In {\em ISWC}, 2017.

\bibitem{BigDatalog}
Alexander Shkapsky, Mohan Yang, Matteo Interlandi, Hsuan Chiu, Tyson Condie,
  and Carlo Zaniolo.
\newblock Big data analytics with datalog queries on spark.
\newblock SIGMOD, 2016.

\bibitem{Storm}
Ankit Toshniwal, Siddarth Taneja, Amit Shukla, Karthik Ramasamy, Jignesh~M.
  Patel, Sanjeev Kulkarni, Jason Jackson, Krishna Gade, Maosong Fu, Jake
  Donham, Nikunj Bhagat, Sailesh Mittal, and Dmitriy Ryaboy.
\newblock Storm@twitter.
\newblock SIGMOD, 2014.

\bibitem{bsp}
Leslie~G. Valiant.
\newblock A bridging model for parallel computation.
\newblock {\em Commun. ACM}, 33(8):103--111, August 1990.

\bibitem{myria}
Jingjing Wang, Magdalena Balazinska, and Daniel Halperin.
\newblock Asynchronous and fault-tolerant recursive datalog evaluation in
  shared-nothing engines.
\newblock {\em {PVLDB}}, 2015.

\bibitem{Spark}
Matei Zaharia, Mosharaf Chowdhury, Tathagata Das, Ankur Dave, Justin Ma, Murphy
  McCauley, Michael~J. Franklin, Scott Shenker, and Ion Stoica.
\newblock Resilient distributed datasets: A fault-tolerant abstraction for
  in-memory cluster computing.
\newblock In {\em NSDI}, 2012.

\bibitem{Sparkstreaming}
Matei Zaharia, Tathagata Das, Haoyuan Li, Timothy Hunter, Scott Shenker, and
  Ion Stoica.
\newblock Discretized streams: Fault-tolerant streaming computation at scale.
\newblock In {\em SOSP}, 2013.

\bibitem{dstreams}
Matei Zaharia, Tathagata Das, Haoyuan Li, Timothy Hunter, Scott Shenker, and
  Ion Stoica.
\newblock Discretized streams: Fault-tolerant streaming computation at scale.
\newblock In {\em Proceedings of the Twenty-Fourth ACM Symposium on Operating
  Systems Principles}, SOSP '13, pages 423--438, New York, NY, USA, 2013. ACM.

\bibitem{SRBench}
Ying Zhang, Pham~Minh Duc, Oscar Corcho, and Jean-Paul Calbimonte.
\newblock Srbench: A streaming rdf/sparql benchmark.
\newblock In {\em ISWC}, 2012.

\end{thebibliography}
\end{document}